# Induction of High-level Behaviors from Problem-solving Traces using Machine Learning Tools

Vivien Robinet, Gilles Bisson, Mirta B. Gordon, Benoît Lemaire

Laboratoire TIMC-IMAG
Faculté de Médecine
38706 La Tronche Cedex, France
first name.last name@imag.fr

**Abstract**: This paper applies machine learning techniques to student modeling. It presents a method for discovering high-level student behaviors from a very large set of low-level traces corresponding to problem-solving actions in a learning environment. Basic actions are encoded into sets of domain-dependent attribute-value patterns called *cases*. Then a domain-independent hierarchical clustering identifies what we call general *attitudes*, yielding automatic diagnosis expressed in natural language, addressed in principle to teachers. The method can be applied to individual students or to entire groups, like a class. We exhibit examples of this system applied to thousands of students' actions in the domain of algebraic transformations.

**Keywords:**. Computer-assisted instruction, Machine learning, Education, Mining methods and algorithms

## 1 Introduction

Many learning environments are able to store very detailed traces of students' activities thus producing huge sets of low-level data. However, identifying high-level behaviors from these data is not straightforward, especially if the concepts of the domain knowledge are not explicitly encoded together with the corresponding traces. In this paper we present a general approach that aims at discovering patterns of student behaviors. Its principles are applicable whenever the information carried by the traces may be split as finite sequences of *{initial state, final state}* pairs, where the final states are the result of basic student transformations performed on the corresponding initial states. Within this context, final states are the initial states of subsequent *{initial state, final state}* pairs (unless they are at the end of the sequence).

Our approach is based on a two-steps procedure:

- a *domain-dependent* representation of the information carried by the traces, which encodes each *{initial state, final state}* pair produced by the student, as a triplet *{context, action, outcome}* that we call a *case*;
- a *domain-independent* machine-learning procedure, based on a clustering technique generating the high-level patterns, that we call *attitudes*.

The output of our system are students' *attitudes*, which are generalizations of the *cases*. They are represented within the same formalism as the *cases*, i.e. *{context, action, outcome}*. Furthermore, *attitudes* are automatically translated into natural language expressions understandable by teachers as well as students themselves. *Attitudes* might be used as inputs to a tutoring system, for instance for generating or selecting a new set of exercises, which may be eventually coupled with the learning environment. Figure 1 displays the general architecture, composed of the learning environment (1), the encoder (2) and the machine learning construction of *attitudes* (3).

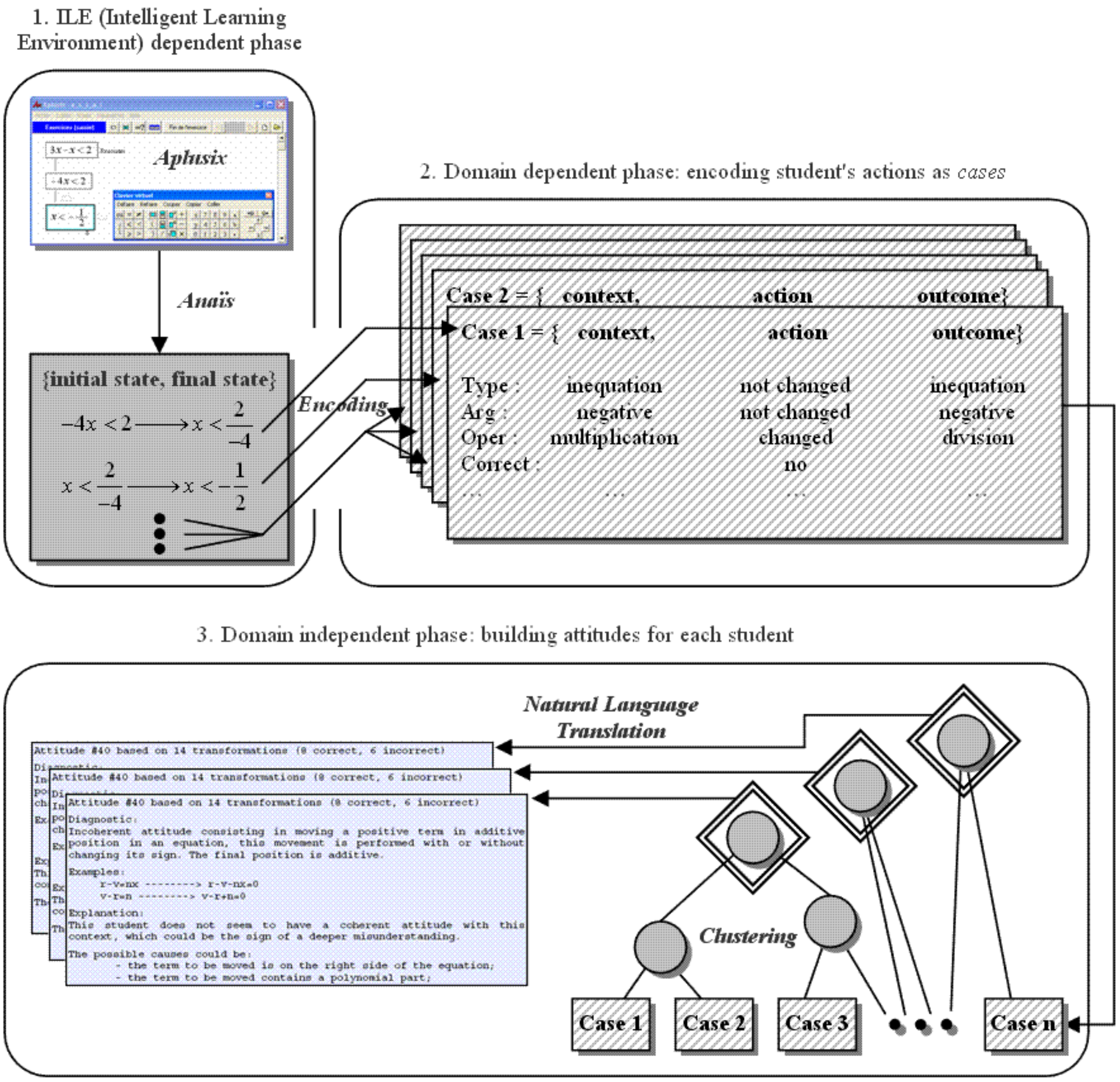

**Fig. 1**. General architecture of our approach

The paper is organized as follows: in section 2 we present the general overview of our approach. The specific learning environment for algebra learning on which we demonstrate our method is presented in section 3. Section 4 presents the domain-dependent encoding procedure and the data representation. The machine-learning procedure and the results are reported on section 5. In section 6 we compare our system to some related work. Finally, we conclude and present possible extensions of our work on section 7.

## 2 General principles of our approach

Our system is intended to be hooked up to a large variety of learning environments that lack an intelligent tracing system. In this section we provide an example to present the general strategy used to identify high-level behaviors, named the *attitudes*, starting with a collection of low-level traces adequately encoded as *cases*.

In the algebra learning context of our present application, a high-level behavior may be, for instance, not modifying the inequation sign when moving a negative multiplicative term from one side of an equation to the other. Since this may arise systematically or just by inattention, we use a statistical approach to assess the significance of local behaviors over a large set of students' *cases*. Our aim is to make relevant generalizations from low-level *case* descriptions to high-level *attitudes*.

Let us show an example of an *attitude* automatically produced by our system from fifty transformations performed by a student. These transformations are mainly movements of terms in equations. A movement is a shortcut which is taught to students (at least in France) to shrink the number of resolution steps. Beginners are taught that to solve an equation such as 7x-4=3, they have to apply the same operation to both sides (adding 4), but later in the studies they are taught that they have to "move" the -4 from one side to the other, while changing its sign. An automatically generated description of an *attitude* produced by our system (typically, our system identifies around 5 to 6 *attitudes* for each student) looks as follows: *Incoherent attitude consisting in moving a positive term in additive position in an equation. This movement is performed with or without changing its sign. The final position is additive.* The expression "with or without" reflects the fact that the system performed a generalization of a sub-part of the student's action ("changing its sign"). Before detailing our method, we first present the algebra learning environment used to collect the student traces.

## 3 Algebra Learning Environment

The APLUSIX learning environment [1] allows students to solve algebraic problems using an equation editor. Given algebraic equations or inequations to be solved, students using APLUSIX proceed step by step as they would do on a notebook. The only imposed constraint is that the expressions entered at any resolution step must be well formed from a syntactic point of view. Figure 2 presents a snapshot of the system, showing a proposed exercise and a student's resolution in three steps. APLUSIX stores all of the student's intermediate results, indicated on the figure as step 1 and step 2 of the resolution. Of course, the granularity of the data continuously varied since the transformation from one student's step to the next one may involve implicit mental operations and/or several simultaneous algebraic transformations. For example, the second step combines two actions: the multiplicative term -4 was moved from the LHS to the denominator of the RHS of the equation without changing the sense of the inequality, and the fraction was then simplified.

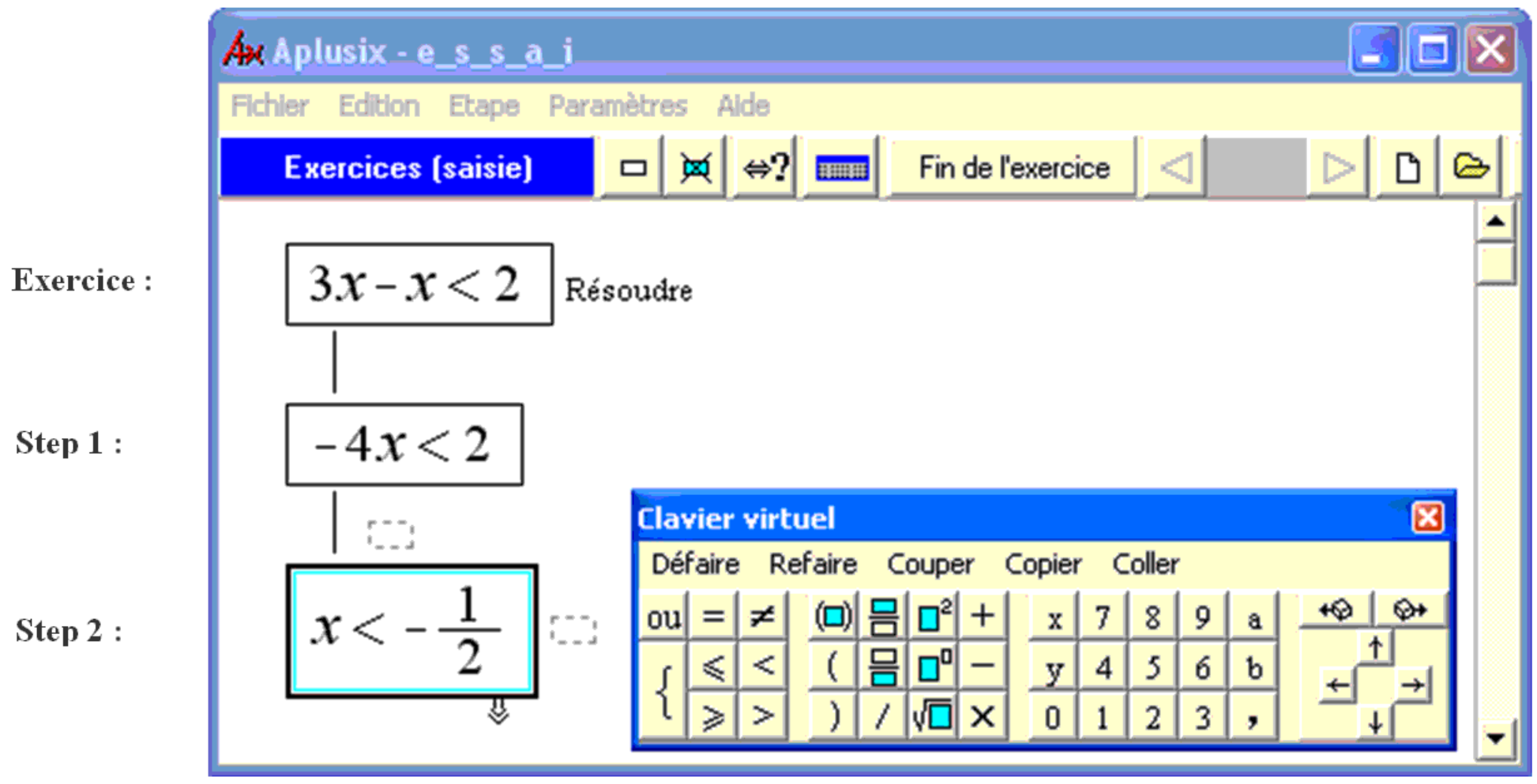

**Fig. 2**. Snapshot of an APLUSIX screen, showing the resolution of an exercise, through a decomposition into 2 steps. Each step may correspond to many elementary algebraic transformations.

In order to implement a systematic treatment and provide an automatic student model, we need to homogenize the granularity of collected data. This is done by introducing whenever necessary virtual elementary steps based on domain knowledge. In the particular case of our algebra learning environment, these steps are produced by ANAIS, a particular software developed by the APLUSIX team, which decomposes the complex student's steps into intermediate elementary steps. To this end, it contains a full set of elementary algebraic rules identified by experimented teachers and didactic experts as being usually implemented by students. These rules are algebraic transformations, that may be either correct (for instance, $(a+b)^2 \rightarrow a^2+2ab+b^2$, or incorrect (for instance $x^n \rightarrow n^x$).

ANAIS strives to describe the student transformations from one step to the next one as resulting from the successive application of rules obtained through a best-first search in the space of all possible algebraic transformations. Accordingly, the student's production is segmented into *{initial state, final state}* pairs, where each final state stems from the corresponding initial state after the application of a single elementary transformation rule. Each pair of states is labeled as correct or incorrect according to the semantic of the rule that has been used to generate it. As a result, we obtain a consistent and homogeneous data set "enriched" with the ANAIS' virtual steps. This set is the input to our system. For example, if a student has performed the following transformation: $-4x < 2 \rightarrow x < -1/2$, ANAIS identifies two steps:

- $-4x < 2 \quad \rightarrow \quad x < 2/(-4)$ (incorrect)
- $x < 2/(-4) \quad \rightarrow \quad x < -1/2$ (correct)

Thus, the two corresponding *{initial state, final state}* pairs are: {-4x < 2, x < 2/(-4))}, labeled incorrect, and {x < 2/(-4), x < -1/2} labeled correct. These are the inputs to our modeling system.

Data used in this paper have been collected in a large scale experiment performed in middle schools in Brazil during the fall 2003-2004. A total number of 2 700 students were asked to solve between 3 and 10 algebraic problems using Aplusix. After segmentation with ANAIS, their production represents 111 258 *{initial state, final state}* pairs, corresponding to an average of 41 resolution steps per student.

## 4 Data Representation: the Domain-dependent Encoder

In this step we transform each domain-dependent *{initial state, final state}* pair into a generic *case*, that is to say a triplet *{context, action, outcome}*, where:

- *context* represents the relevant part of the initial state with respect to the semantics of the transformation performed by the student;
- *action* represents the action itself, based on an automatic analysis of the differences between initial state and final state and using the correctness label described above;
- *outcome* represents the relevant part of the final state.

Each item of the *case triplet* is encoded as a set of attributes, in order to find regularities and identify general behaviors using machine-learning approaches. We organize the attributes describing these components into three categories, reflecting three different levels of location in the initial and final states. These categories are:

- *argument*: descriptors of the element(s) directly concerned by the transformation between the initial and the final state;
- *local term* (simply called term hereafter): descriptors of the elements that are close to the argument;
- *expression: descriptors of global properties of the state.*

Coming from our domain of algebra learning, here is an example (Figure 3) where we show the three descriptor levels of the initial state of an incorrect transformation.

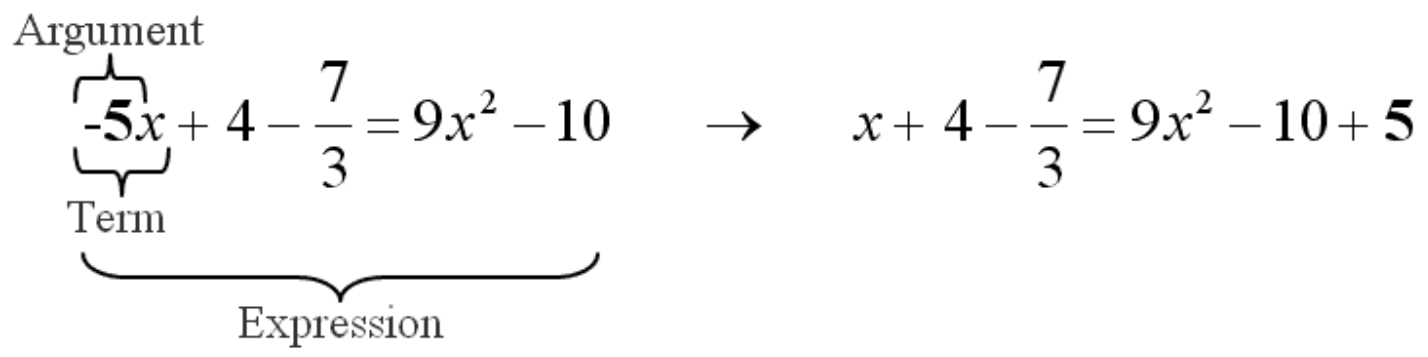

**Fig. 3**. The three different levels of location: argument, term and expression

In the domain of movements in algebraic equations, we defined 25 attributes to express the context, 6 for the action and 6 for the outcome, totalizing 37 attributes for each *case*. Depending on their nature, attributes may take different discrete values. Table 1 contains the list of the most relevant ones for the transformation described in Figure 3. Note that the non-relevant attributes are not detailed here: being unchanged by the action, they remain thus identical in the context and the outcome.

**Table. 1**. Representation of a *case* in the *{context, action, outcome}* formalism

| Context | | Action | | Outcome | |
|---|---|---|---|---|---|
| arg.side | *left* | | | | |
| arg.location | *beginning* | | | | |
| arg.polynomial | *false* | | | | |
| arg.coefficient | *true* | | | | |
| arg.implicitSign | *false* | | | | |
| arg.operateur | × | arg.operateurChanged | *true* | arg.operateur | + |
| arg.category | *multiplicat* | arg.categoryChanged | *true* | arg.category | *add* |
| arg.negative | *true* | arg.signChanged | *true* | arg.negative | *false* |
| term.polynomial | *true* | | | | |
| expr.type | *equation* | expr.typeChanged | *false* | expr.type | *equation* |
| expr.polynomial | *true* | | | | |
| | | expr.correct | *false* | | |

The content of the table can be interpreted as follow: the context attributes say that the argument is in the left hand side of the equation, at the beginning, it is not polynomial, it is an integer with an explicit negative sign and the operator is multiplicative. The term is polynomial. The expression is a polynomial equation. The outcome attributes say that, after the transformation the operator of the argument is an addition, it belongs to an additive category and is positive. The expression is still an equation. The action attributes are derived from the context and outcome attributes. They indicate that the operator of the argument has been changed by the student, that its category and its sign have also been modified. However, the type of the expression remains the same. In addition, the last attribute indicates that this transformation is algebraically incorrect.

It is worth noting that some of these attributes are redundant in this example, but they are needed to describe other students' behaviors. The aim is to use attributes that allow the model to give a fine explanation of students' behaviors, even if some of them are redundant. The generalization process will select which of them best explain the transformation.

Provided the {initial state, final state} pairs are represented by *cases* of {context, action, outcome} triplets, our approach can be fruitfully used to provide behavioral *attitudes* of the students. *Attitudes* are generalizations of student's *cases*, performed by an independent module we will present in the next section. Its role is to identify high-level behavior from this low-level data.

# 5 Discovering *Attitudes*

## 5.1 Technique: Hierarchical Clustering

*Cases* are the basic material used by our system to uncover high-level student's behavior. Our approach relies on an unsupervised learning algorithm to cluster similar *cases* into classes hereafter called *attitudes*. The goal is to get a set of a few classes, representative of typical student's behavior. We use a hierarchical clustering technique [2]. This algorithm groups together the two most similar (according to a distance detailed below cf. 5.3) *cases* into a *working cluster* that replaces the corresponding *cases*. The procedure is applied again and again on the set of remaining *cases* and *working clusters*. The latter are candidate *attitudes*: they generalize the underlying *cases*. The algorithm stops when the closest similarity between elements reaches a given threshold.

Our attribute-based representation is combined with a statistical counting that keeps trace of the number of *cases* that share the same attribute value in the *working cluster* or *attitude*. We keep track of this statistical information to characterize the way attributes are generalized, and whether this generalization is significant or not as we explained in part 2, the goal being to distinguish between systematic or occasional student's actions. *Cases* have one and only one counter set to 1 for each attribute, the one corresponding to the actual value of the attribute. When we group together two *cases* or *working clusters*, the counters of the attribute values are updated. Table 2 contains an example in which *case* 12 is grouped with *working cluster* 6, giving a new *working cluster* that generalizes (and replaces) both of them. The attribute values in the *attitudes* represent the numbers of *cases* sharing the corresponding attribute value in the cluster.

**Table. 2**. Generalization of one *case* and one *working cluster* producing a new *working cluster*

| Attributes | Case 12 | | | | | Working Cluster 6 | | | | | Working Cluster (6&12) | | | |
|---|---|---|---|---|---|---|---|---|---|---|---|---|---|---|
| arg.side | left | right | | | + | left | right | | | → | left | right | | |
| | 0 | 1 | | | | 1 | 3 | | | | 1 | 4 | | |
| arg.location | beg. | mid. | end | alone | | beg. | mid. | end | alone | | beg. | mid. | end | alone |
| | 0 | 1 | 0 | 0 | | 0 | 1 | 3 | 0 | | 0 | 2 | 3 | 0 |
| arg.complex | true | false | | | | true | false | | | | true | false | | |
| | 1 | 0 | | | | 0 | 1 | | | | 1 | 1 | | |
| arg.polynomial | true | false | | | | true | false | | | | true | false | | |
| | 1 | 0 | | | | 3 | 0 | | | | 4 | 0 | | |
| … | … | … | | | | … | … | | | | … | … | | |

## 5.2 Different kinds of *attitudes*

An *attitude* is a generalization of underlying *cases*. The attribute "expr.correct" has a particular meaning. It is not used during the generalization process, but is very important to characterize the *attitudes* obtained. We distinguish two kinds of *attitudes*:

- *coherent attitudes*, that are either correct or incorrect;
- *incoherent attitudes* which contain a statistically significant proportion of correct and incorrect *cases*. The fact that both proportions are significantly not equal to zero implies that it is probably not an isolated case but rather a more systematic behavior.

## 5.3 The distance

To compare the pairs of *cases* or *working clusters* we use a distance index taking into account the differences between the context part, the action part and the outcome part of the two *case* triplets considered. This distance relies on a coefficient α emphasizing the context part or the action/outcome part (cf. Equation 1). If the context part is given more weight, the algorithm tends not to cluster *attitudes* that have distant contexts. The system is then more likely to discover incoherent behaviors (i.e. attitudes in which the student performs different actions in similar contexts). In the other way, if the action and outcome parts are given more weight, the system does not tend to group *attitudes* with distant actions, even if contexts are similar. This would lead to the discovery of coherent behaviors (i.e. attitudes in which the student performs similar actions in different contexts). The general distance $D_{Att}$ between two *cases* or *attitudes* $A_i$ and $A_j$ is the following (ctx, act and out stand for context, action and outcome):

$$D_{Att}\left(A_i, A_j\right) = \alpha \times dist\left(ctx\left(A_i\right), ctx\left(A_j\right)\right) + (1-\alpha) \times \left(dist\left(act\left(A_i\right), act\left(A_j\right)\right) + dist\left(out\left(A_i\right), out\left(A_j\right)\right)\right) \qquad (Eq.1)$$

where the distance *dist* between sets of attributes of a given category, depends on the frequency of values for each attributes. Each attribute *d* is weighted by an integer $p_d$ *:*

$$dist\left(V_i, V_j\right) = \sum_d \left( p_d \times \sum_m \left| \frac{\#V_{i,d,m}}{\sum V_{i,d}} - \frac{\#V_{j,d,m}}{\sum V_{j,d}} \right| \right) \quad \text{with} \begin{cases} \#V_{i,d,m} & \text{: frequency of value } m \text{ for attribute } d \\ \sum V_{i,d} & \text{: number of accurrences of attribute } d \end{cases} \qquad (Eq.2)$$

Now we are going to explain how the clustering can be used to analyze not only the individual behavior of each student (cf. 5.4) but also to provide a snapshot of the behavior of a group of student (cf. 5.5).

## 5.4 Individual *Attitudes*

We applied this method on 111 258 transformations collected from 2 700 students in Brazil. Figure 4 displays the hierarchical clustering of all the transformations produced by student #1497 based on fifty transformations. The full tree (Figure 4) is shown for illustrative purposes. The algorithm was actually stopped at the dashed line that represents the chosen generality level, which corresponds to a similarity threshold of 0.38. This threshold appears to be a good value according to the conducted tests. We use colors and shapes of the nodes to represent the most relevant attributes.

- The color indicates the correctness of the node (correct: light gray, incorrect: dark gray, both: middle gray).
- The shape represents the operator (+: triangle, -: square, *: pentagon, /: circle).

Let us describe some of the five *attitudes* obtained (with our threshold) in this example.

- *Attitude* #1 corresponds to a *correct and coherent* behavior. Its attributes indicate that the student knows how to solve simple equations (where "argument.squareRoot", "argument.power" and "argument.fraction" are false) in which a negative term ("argument.negative"=true) has to be moved. It is the case of transformations like : 6x**-3**=2x+4 → 6x=2x+4**+3** where the *argument* is represented in bold. The student correctly moves the argument to the other side, whatever its position ("argument.side" and "argument.position" are generalized) or its coefficient ("arg.coefficient" is generalized), the argument of the outcome is still in additive position ("argument.categoryChanged"=false), but the sign has changed ("argument.signChanged"=true).
- *Attitude* #5 is an example of an *incoherent attitude*. In a similar case (simple equations), but with a positive argument ("argument.negative"=false), the student sometimes fails to change the sign of the argument.

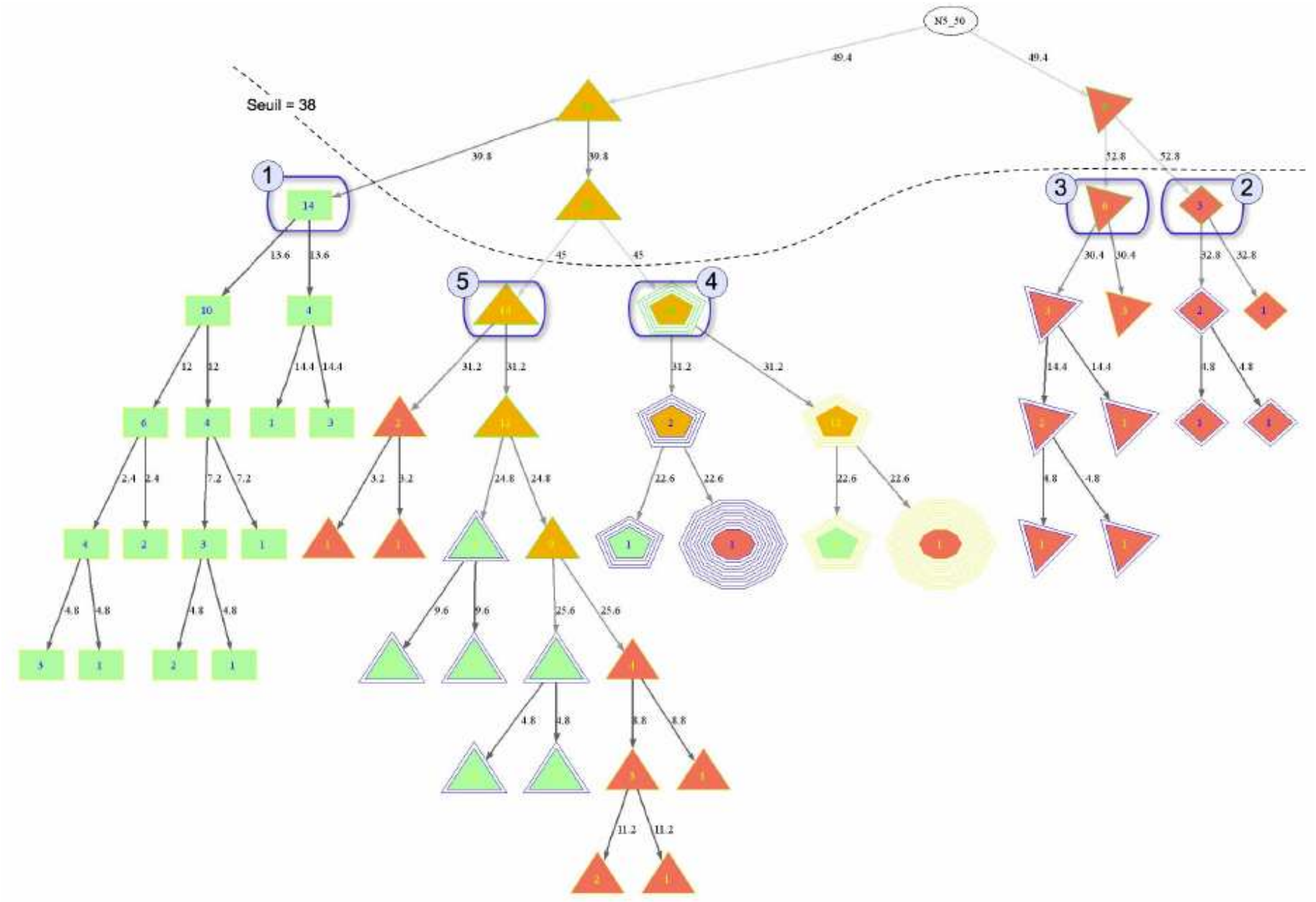

**Fig.4**. Hierarchical clustering of *cases* for student #1497. Five *attitudes* have been kept. Leaves contain one or more identical *cases*.

The aim of discovering *attitudes* is mainly to allow teachers to obtain a precise diagnosis about students. In order to produce a more legible diagnosis of each student's production, we transform the *attitudes*' attribute values into a natural language text (Figure 5), by concatenating predefined sentences. We also automatically generate two examples and a small comment about the coherence or incoherence for each *attitude*. Whenever the *attitude* is incoherent, an algorithm goes back down through the hierarchical tree until reaching the first node that clustered two coherent *attitudes*. It then looks for attributes that discriminate between both *attitudes*. These attributes are also provided because they may be correlated with the reason for the student's incoherent behavior. Here is an example of such a diagnosis generated automatically by our system:

```
Attitude #40 based on 14 transformations (8 correct, 6 incorrect)
Diagnostic:
Incoherent attitude consisting in moving a positive term in additive position in
an equation, this movement is performed with or without changing its sign. The
final position is additive.
Examples:
     r-v=nx --------> r-v-nx=0
     v-r=n --------> v-r+n=0
Explanation:
This student does not seem to have a coherent attitude with this context, which
could be the sign of a deeper misunderstanding.
The possible causes could be:
       - the term to be moved is on the right side of the equation;
       - the term to be moved contains a polynomial part;
```

**Fig.5**. Natural language translation of an *attitude* of student #1497

Another usage of our automatic *attitude* discovery, currently under investigation, is to automatically generate appropriate exercises for students in case of incorrect or incoherent *attitudes*. For instance, the above mentioned *attitude* #5 would lead to the generation of an exercise in which a positive term has to be moved to the other side of an equation. For example: 7x+4=11x+13.

### 5.5 Group *Attitudes*

Processing all students' *cases* produced 11 026 *attitudes*. Their global analysis identified whether some of these *attitudes* were shared by several students. It is not possible to simply draw a frequency chart because very few *attitudes* are fully identical among different students, since they are the result of an induction process. It is thus necessary to aggregate individual *attitudes*. To this end, we use the same mechanism as before because individual *cases* and *attitudes* share the same formalism. The similarity threshold was set to a low value (0.1) because the goal is not to generalize *attitudes* but rather to smooth the differences between individual *attitudes*. Figure 6 displays the 38 most frequent *attitudes*. The y-axis indicates the number of individual *cases* that compose each *attitude*, together with their correctness. The number of students is also displayed.

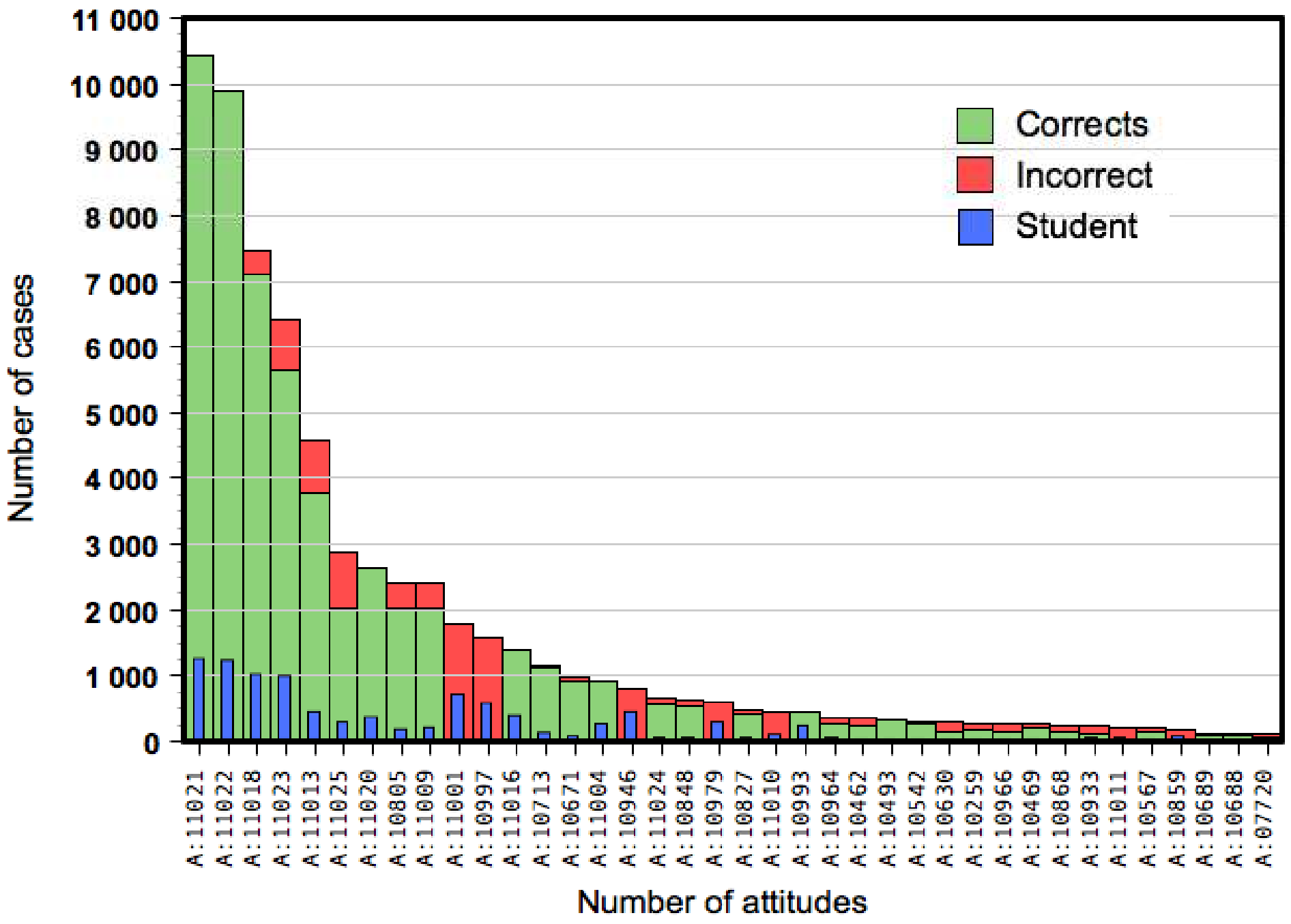

**Fig. 6**. Histogram of the 38 most frequent *attitudes*. Green and red bars represent numbers of correct and incorrect *cases*; the blue bars are the number of students presenting the corresponding *cases* in their productions.

In our data, the two most frequent *attitudes* are correct ones. They correspond to a movement of a positive argument (*attitude* #11021, 1217 students) or a negative argument (*attitude* #11022, 1188 students). Incorrect *attitudes* have also been identified. For instance, *attitude* #10997 (565 students) is an additive movement of a negative argument from one side to the other of an inequation, without changing its sign. There are also incoherent *attitudes*: *attitude* #11023 (955 students) is an additive movement in an inequation in which the sign of a negative argument is correctly changed, but the inequation sign is sometimes also reversed, probably because of a confusion with multiplicative arguments.

## 6 Related work

Students' data produced by interactive learning environments are quite often huge sequences of low-level descriptions which should be automatically interpreted by changing the level of granularity [3]. Several existing systems rely on machine learning techniques to discover student knowledge behind such basic descriptions. Extracting regularities requires a rewriting of student's productions in term of higher level domain-dependent attributes defined by experts.

Many systems build user models by means of supervised machine learning techniques based on predefined profiles provided by domain experts. Profile Extractor [4] induces rules from pre-classified examples, using a decision tree. Its goal is to discover preferences, needs and interests of e-learning students. Our approach is quite different since our goal is to automatically discover those profiles.

Animalwatch [5] is closer to our system. Its domain is basic arithmetic. Animalwatch analyzes a student data to predict whether she would be able to solve the problem and the time it would take her. Animalwatch uses four kinds of variables, similar to our attributes:

- *Student*: student gender, scores to initial tests;
- *Topic*: type of operator (addition, substraction, etc.) and type of operands;
- *Problem*: problem difficulty, number and difficulty of prerequisites to solve the problem (such as adding fractions, simplifying fractions), etc.
- *Context*: number of prior errors, best hint seen, etc.

After tests of several algorithms, such as a Bayesian classifier and a decision tree, the authors finally use a simple linear regression to predict the two variables. The main difference between Animalwatch and our system is that we are not attempting to predict performances but rather to construct a cognitive profile of the student.

Web-EasyMath [6] also relies on machine learning algorithms to construct student models in the domain of algebraic powers. The goal is to define at best a model for a new user. The student is first required to pass a test about her knowledge of the four basic operations and to assess her self-estimation on basic skills. A distance weighted k-nearest neighbor algorithm is used to asses the concept knowledge level of the new student with respect to all the students that belong to the same category.

With a more generic scope, Sison & Shimura [7] propose several features that might be used to categorize systems that discover student knowledge from their behaviors. Let us define our approach with respect to some of these features:

- *student behavior complexity (*from simple values to more complex expressions). The student behavior is undoubtedly complex in our system;
- *student behavior multiplicity (*from single behavior to multiple behaviors) We are not analyzing in depth a single behavior, our system rather considers a very large set of behaviors;
- *background knowledge construction* (from completely specified to automatically extended). In our case, the domain knowledge, either correct or incorrect, cannot be extended by the system itself.
- *student model construction* (analytic or synthetic). Our approach is synthetic because it is based on behavior generalization. However, the ANAIS software which attempts to discover intermediate resolution steps is analytic.

Finally, our system can be analyzed with respect to Mayo & Mitrovic's classification [8]. They proposed a threefold classification of existing intelligent tutoring systems:

- *expert-centric systems* in which the internal representation of the domain is mainly designed by an expert;
- *efficiency-centric systems* which are partially specified and contain parameters that allow to optimize a certain criterion (evaluation time, memory used, etc.);
- *data-centric systems* which learn their structure using mainly data.

This classification was initially specific to Bayesian student modeling, but could be easily extended to other approaches. In our case, an *attitude* in not a pre-defined expert object, but is constructed by a generalization process using data produced by the student. Our approach could therefore be considered in this classification as a data-centric student modeling approach.

## 7 Conclusion

This paper presents a system allowing to automatically uncovering high-level *attitudes* of students out of problem-solving traces produced in a learning environment. Our general purpose approach makes the system applicable to many learning domains, under the assumptions that the student actions can be represented as (context, action, outcome) triplets. The system's output is a synthesis, directly understandable by teachers or didactic experts, of the knowledge of a student or a class. The system can deal with incoherent behaviors and distinguish between occasional or systematic student errors. The results may be used for automatically generating new appropriate exercises.

The domain on which we applied our system is that of algebraic transformations, mainly additive and multiplicative movements in equations and inequations. Applications to factoring and reducing algebraic expressions are currently in progress.

Modeling student actions by means of a set of attributes is an important feature of our approach. We could have used other formalisms. For instance, in our algebra domain, student actions could have been pairs of equations represented as trees and we could have invented formalisms for representing generalized actions. However, this formalism would have been too much dependent on our domain and would not have been easily extended to other domains. Attributes are a much more general way of representing student actions, especially at the low level from which our approach can perform generalizations. This formalism allows a clear distinction between the domain knowledge and the machine-learning process of building the student's model. Attributes are obviously domain-dependent, but once they have been defined, the machine-learning mechanism is ready to operate. As a consequence, the diagnosis will be expressed in terms of the attributes, thus understandable by humans.

It is worth noting that attributes do not have to be cleverly designed in order to be independent from each other: as we mentioned earlier, the generalization process will automatically select those which best explain the student behavior, provided there are enough examples. Our system is based on the hypothesis that student traces are temporal sequences of states, which we know is not the case for every domain. Going from one state to the other is done by only one action, the cause of a state being the only preceding state. This is probably our strongest hypothesis, but we believe that many problem-solving learning environments are based on this hypothesis.

Another limit of our approach is that it does not take into account the order in which the student is exposed to exercises. This information may be very useful to model the time course of learning, through the analysis of which *attitudes* appear or disappear on time. One approach could be to rely on a incremental clustering system such as Cobweb [9]. The information about student steps order are needed if we want to understand the student resolution strategies. It may certainly give richer diagnosis.

## Acknowledgements

It is a pleasure to thank David Renaudie [10] as well as all members of the Did@tic team of the Leibniz Laboratory for their interest in this work, and fruitful discussions. This research has been in part supported by a *Cognitique* grant from the French Ministry of Research.